\documentclass[sigconf]{acmart}
\usepackage{stfloats}
\AtBeginDocument{%
  \providecommand\BibTeX{{%
    \normalfont B\kern-0.5em{\scshape i\kern-0.25em b}\kern-0.8em\TeX}}}

\setcopyright{iw3c2w3}

\begin{document}

\copyrightyear{2021}
\acmYear{2021}
\acmConference[WWW '21]{Proceedings of the Web Conference 2021}{April 19--23, 2021}{Ljubljana, Slovenia}
\acmBooktitle{Proceedings of the Web Conference 2021 (WWW '21), April 19--23, 2021, Ljubljana, Slovenia}
\acmPrice{}
\acmDOI{10.1145/3442381.3449929}
\acmISBN{978-1-4503-8312-7/21/04}

\title{Soft-mask: Adaptive Substructure Extractions for Graph Neural Networks}

\author{Mingqi Yang}
\affiliation{
\institution{Dalian University of Technology}
\state{Dalian}
\country{China}
}
\email{yangmq@mail.dlut.edu.cn}

\author{Yanming Shen}
\authornote{Corresponding author}
\affiliation{
\institution{Dalian University of Technology}
\state{Dalian}
\country{China}
}
\email{shen@dlut.edu.cn}

\author{Heng Qi}
\affiliation{
\institution{Dalian University of Technology}
\state{Dalian}
\country{China}
}
\email{hengqi@dlut.edu.cn}

\author{Baocai Yin}
\affiliation{
\institution{Dalian University of Technology}
\institution{Peng Cheng Laboratory }
\state{Dalian}
\country{China}
}
\email{ybc@dlut.edu.cn}

\renewcommand{\shortauthors}{M. Yang and Y. Shen, et al.}

\begin{abstract}
For learning graph representations, not all detailed structures within a graph are relevant to the given graph tasks. Task-relevant structures can be $localized$ or $sparse$ which are only involved in subgraphs or characterized by the interactions of subgraphs (a hierarchical perspective). A graph neural network should be able to efficiently extract task-relevant structures and be invariant to irrelevant parts, which is challenging for general message passing GNNs. In this work, we propose to learn graph representations from a sequence of subgraphs of the original graph to better capture task-relevant substructures or hierarchical structures and skip $noisy$ parts. To this end, we design soft-mask GNN layer to extract desired subgraphs through the mask mechanism. The soft-mask is defined in a continuous space to maintain the differentiability and characterize the weights of different parts. Compared with existing subgraph or hierarchical representation learning methods and graph pooling operations, the soft-mask GNN layer is not limited by the fixed sample or drop ratio, and therefore is more flexible to extract subgraphs with arbitrary sizes. Extensive experiments on public graph benchmarks show that soft-mask mechanism brings performance improvements. And it also provides interpretability where visualizing the values of masks in each layer allows us to have an insight into the structures learned by the model.
\end{abstract}



\maketitle
\section{Introduction}

Graph structure data is ubiquitous.
Molecules, social networks, and many other applications can be modeled as graphs.
Recently, graph neural networks (GNNs) have shown their power in graph representation learning \cite{kipf2016semi,hamilton2017inductive,bronstein2017geometric}.
General GNNs follow neighborhood aggregation (or message passing) scheme \cite{gilmer2017neural}, where node representations are computed iteratively by aggregating transformed representations of its neighbors.
The aggregation operation on each node is shared with the same parameters, with structural information learned implicitly.

One issue of this kind of scheme is that task-relevant structural information is likely to be mixed up with irrelevant (or noisy) parts, making it indistinguishable for the downstream processing, especially for long-range dependency captured by higher layers in a deep model \cite{li2018deeper,chen2020measuring,oono2019graph}.
To handle this, one possible strategy is to improve the ability to distinguish different graph topologies.
Hopefully, useful structural information will be retained and passed to the higher layers for further processing.
This strategy is followed by GIN \cite{xu2018how} whose ability to distinguish different graph structures is equivalent to 1-order Weisfeiler-Lehman (WL) graph isomorphism test.
Other GNN implementations \cite{morris2019weisfeiler,chen2019equivalence,maron2019provably} with discrimination power in analogy to high-order WL test are also proposed.

Ideally, GNNs should behave in an $extract$ manner.
For noisy graphs, the learned representations should correspond to the subgraph with noisy parts not involved.
To achieve this, one basic idea is to restrict all layers learing on the same subgraph.
It can be considered that a $k$-layer GNN learns on a sequence of subgraphs of length $k$.

\begin{figure}[h]
\centering
\includegraphics[width=0.85\linewidth]{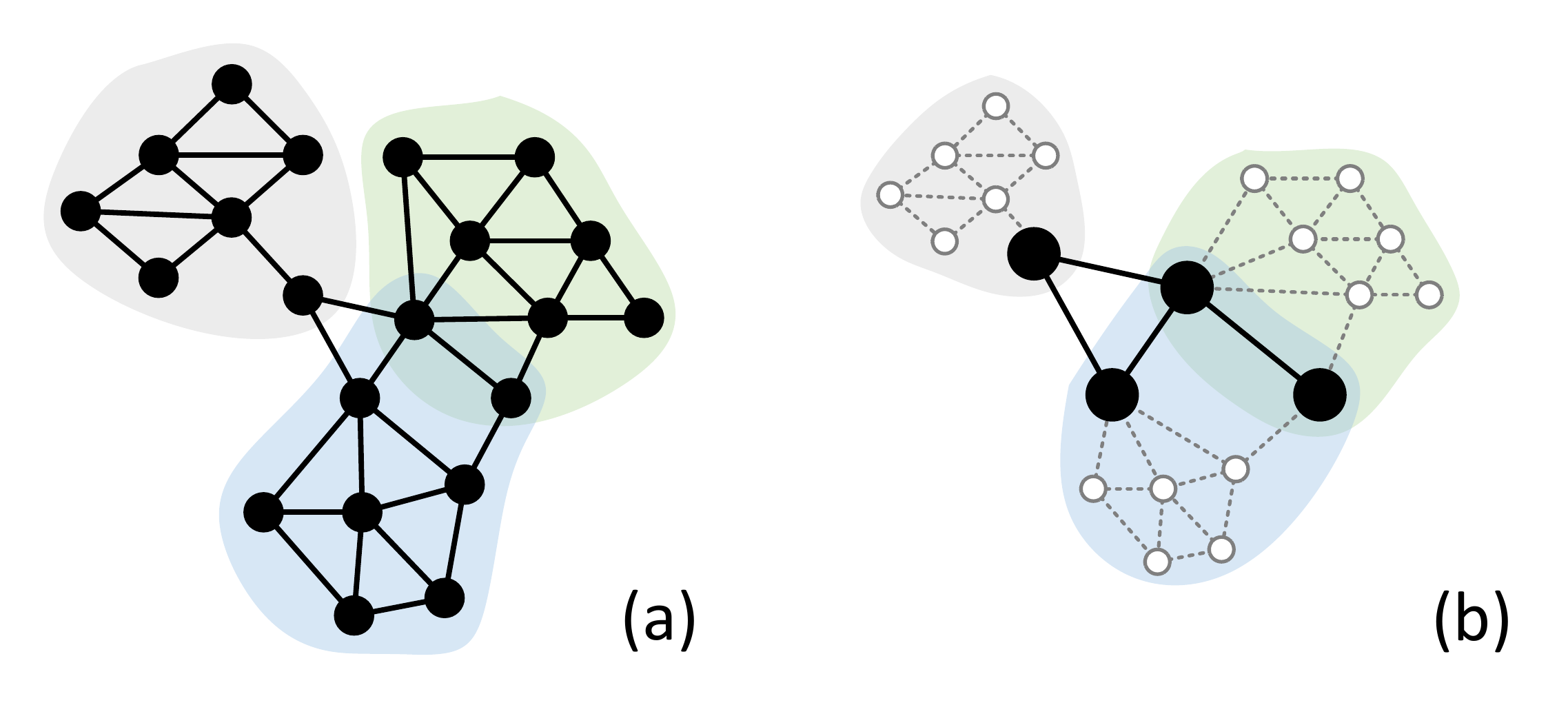}
\caption{An example of hierarchical graph structures.}
\label{Intro:hierarchy}
\end{figure}
For graphs characterized with hierarchical structures,
to explicitly capture the hierarchical structures,
the neural network should also encode them in a hierarchical manner.
We explain this motivation empirically with a simple example of hierarchical graphs with a height of 2 as given in Figure \ref{Intro:hierarchy} (a),
where the structure is identified by three individual parts (in different colors) and their interactions.
Figure \ref{Intro:hierarchy} (b) demonstrates that in this graph, nodes can be divided into two groups:
\begin{itemize}
\item \textbf{Leaf node.}
Nodes only have connections within each part such as white nodes in Figure \ref{Intro:hierarchy}(b);
\item \textbf{Root node.}
Nodes have connections with other parts or shared with other parts such as black nodes in Figure \ref{Intro:hierarchy}(b). 
\end{itemize}
Clearly,
the interactions among different parts are only decided by root nodes.
By removing all leaf nodes,
we obtain a $reduced$ subgraph that only demonstrates the interactions among different parts.
To explicitly capture the interactions among different parts,
the aggregation in lower layers should be only conducted on leaf nodes to encodes each part individually.
Then root nodes are taken into consideration to encode the interactions of each part in higher layers.
This requires that each layer of a neural network should be flexible to learn on any given subgraphs,
and correspondingly the final representations are learned from a sequence of subgraphs.
Since hierarchical graph structures can be viewed recursively as each individual part and their interactions,
this mechanism is natural to be extended to hierarchical graphs with heights larger than 2.

According to the above analysis,
learning graph representations from a sequence of subgraphs actually unifies subgraph representation learning and hierarchical representation learning.
To reach this goal, there are two main challenges:
\begin{itemize}
\item
How to efficiently represent any random subgraph in each layer while maintaining differentiability?
\item
How to decide the sequence of subgraphs learned by the model?
\end{itemize}

GAM \cite{lee2018graph} extracts desired substructures by representing a graph with a fixed-length sequence of nodes sampled from the original graph.
Top-$k$ graph poolings \cite{gao2019graph,lee2019self,knyazev2019understanding} score the importance of each node and drop low score nodes and related edges layer by layer.
These strategies can dynamically focus on informative parts, making the neural network more robust in dealing with noisy graphs.
However, GAM requires the sampled node sequence to be with the fixed length, and top-$k$ poolings require a fixed drop ratio.
These limitations can be obstacles to represent desired subgraphs with arbitrary scales.

To address these issues,
we propose soft-mask GNN (SMG) layer.
It extracts subgraphs of any size by controlling node masks.
Then the problem of finding the desired subgraph is converted to finding proper mask assignments.
We theoretically show that the learned representation by a soft-mask GNN layer with proper mask assignments is equivalent to the corresponding subgraph representation.
Stacking multiple such layers leads to the learned representation of a sequence of subgraphs that can be used to capture the $informative$ substructures as well as hierarchical structures.
Different from general GNNs that follow $dense$ aggregation where the aggregation is conducted on all nodes, the aggregation in a soft-mask GNN layer can adaptively $shutdown$ the aggregation of undesired parts while maintaining differentiability, which we call $sparse$ aggregation.

The soft-mask GNN layer applies continuous mask values in order to maintain the differentiability of the networks.
It characterizes the weights of different parts within a graph, which can also be considered as a global attention mechanism.
This global attention mechanism provides interpretability,
where visualizing mask value distributions in different layers on some public graph benchmarks provides insights of informative parts or hierarchical structures learned by the model.

Our contributions are summarized as follows:
\begin{itemize}
\item We propose the soft-mask GNN layer which achieves a kind of sparse aggregation in general GNNs.
It is used to represent any given subgraph with an arbitrary size.
\item We theoretically analyze that by learning a graph representation from a sequence of individual subgraphs of the original graph, our model is capable of extracting any desired substructures or hierarchical structures.
\item We evaluate the soft-mask GNN on public graph benchmarks and show a significant improvement over state-of-the-art approaches. Furthermore, by visualizing the mask values in different layers, we provide insights on structural information learned by the model.
\end{itemize}

\section{Preliminaries}

\subsection{Notations}

For a graph $G$, we denote the set of edges, nodes and node feature vectors respectively by $E_{G}$, $V_{G}$ and $X_{G}$.
$\mathcal{N}(v)$ is the set of $neighbors$ of node $v$, i.e., $\mathcal{N}(v)=\{u\in V_{G}|(u,v)\in E_{G}\}$. Let $\mathcal{\hat N}(v)=\{v\}\cup\mathcal{N}(v)$.
Let $S$ be a subset of nodes, i.e., $S\subseteq V_{G}$. Then the subgraph $G_{S}$ of $G$ is the graph whose vertex set is $S$ and whose edge set consists of all of the edges in $E_{G}$ that have both endpoints in $S$.
Also, we use $[n]$ to denote $\{1,2,...,n\}$ and $\{\{...\}\}$ to denote a multiset, i.e., a set with possibly repeating elements.

\subsection{Graph Neural Networks}

GNNs learn the representation vector of a node, $\mathbf h_{v}$, or the entire graph, $\mathbf h_{G}$, from a graph, utilizing both node features and the structure of the graph.
Most proposed GNNs fit within the neighborhood aggregation framework \cite{gilmer2017neural}. Formally, the $k$-th layer of a GNN is
\begin{equation}
\begin{aligned}
\mathbf h^{(k)}_{v}=\textrm{Update}(\mathbf h^{(k-1)}_{v}, \textrm{Aggregate}(\{\{\mathbf h^{(k-1)}_{u}|u\in\mathcal N(v)\}\})).
\end{aligned}
\label{mpnn}
\end{equation}
The node representations from the last iteration are then passed to a classier for node classification. For graph-level tasks such as graph classification, the $\textrm{Readout}$ function is needed which aggregates all node representations from the last iteration to obtain the entire graph representation
\begin{equation}
\begin{aligned}
\mathbf h_{G}=\textrm{Readout}(\{\{\mathbf h^{(K)}_{v}|v\in V_{G}\}\}),
\end{aligned}
\label{readout}
\end{equation}
where $\mathbf h^{(K)}_{v}$ is node representations learned by a $K$-layer GNN. $\mathbf h_{G}$ is then passed to a classifier for graph classification.

Aggregate(.) in Equation \ref{mpnn} is used to aggregate neighbors' representations to generate current node representation, and Readout(.) in Equation \ref{readout} is used to aggregate all node representations to generate the entire graph representation.
In order for the GNNs to be invariant to graph isomorphism, Aggregate(.) and Readout(.) should be permutation invariant.
Meanwhile, Equation \ref{mpnn} and Equation \ref{readout} should be differentiable to make the network trainable through backpropagation.

\section{Soft-mask GNN}

In this section,
we first present the SMG layer and explain how it is used to extract the desired subgraph by controlling the mask assignments.
Then, we show how a multi-layer SMG learns subgraph representations and hierarchical representations respectively with corresponding subgraph sequences.
We also present a method to compute mask assignments, which can automatically extract substructures or hierarchical structures through backpropagation.
Finally, we generalize the SMG to multi-channel scenarios.

\subsection{Sparse Aggregation and Subgraph Representation Learning}
\label{sparse-subgraph}

GNN operation is viewed as a kind of smoothing operation \cite{li2018deeper,chen2020measuring,oono2019graph}, where each node exchanges feature information with its neighbors.
This kind of scheme does not skip irrelevant parts explicitly.
On the contrary, our SMG layer extracts substructures as follows:
\begin{equation}
\begin{aligned}
\mathbf h^{(k)}_{v}=\textrm{ReLU}(\mathbf W^{(k)}_{1}m^{(k)}_{v}[\mathbf h^{(k-1)}_{v}||\sum_{u\in\mathcal N(v)}m^{(k)}_{u}\mathbf h^{(k-1)}_{u}]),
\end{aligned}
\label{mask-gnn}
\end{equation}
where $m^{(k)}_{v}\in[0,1]$ refers to the soft-mask of node $v$ at the $k$-th layer,
$||$ denotes concatenation operation,
and $\mathbf W^{(k)}_{1}\in\mathbf R^{d\times d}$ is a trainable matrix, where $d$ is the dimension of node representations.
The computation of $m^{(k)}_{v}$ will be presented in Section \ref{soft-mask}.
The SMG layer should satisfy the following constraints:
\begin{itemize}
\item The linear transformation $\mathbf W^{(k)}_{1}$ does not include constant part (i.e., bias);
\item The activation function (ReLU) satisfies $\sigma(0)=0$;
\item The aggregation operator (SUM) is $zero$ $invariant$, which is formally defined as follows:
\end{itemize}
\textbf{Zero invariant.} 
The aggregation function $f:\{\{\mathbb R^{m}\}\}\rightarrow \mathbb R^{n}$ is zero invariant if and only if
\begin{equation}
\begin{aligned}
\nonumber
f(\mathcal S)=
\begin{cases}
\mathbf0&\mathcal S=\emptyset \\
f(\mathcal S/\{\mathbf0\})&\text{otherwise},
\end{cases}
\end{aligned}
\label{zero-invariant}
\end{equation}
where $\mathcal S$ is a multiset of vectors.
Note that the SUM operator is zero invariant while the MEAN is not.

With the above restrictions, setting $m^{(k)}_{v}=0$ leads to $\mathbf h^{(k)}_{v}=\mathbf 0$ and
$\mathbf h^{(k)}_{u}=\textrm{ReLU}(\mathbf W^{(k)}_{1}m^{(k)}_{u}[\mathbf h^{(k-1)}_{u}||\sum_{u^{\prime}\in\mathcal{N}(u)/\{v\}}m^{(k)}_{u^{\prime}}\mathbf h^{(k-1)}_{u^{\prime}}])$
for any $u\in\mathcal{N}(v)$, which is said that node $v$ is inaccessible for its neighbors.
Therefore, for any subgraph $G_{S}$ of $G$,
the $1$-layer SMG together with a zero invariant Readout function can represent $G_{S}$ and completely skip other parts by controlling the mask assignments as follows:
\begin{equation}
\nonumber
\begin{aligned}
\mathbf h_G|&_{m^{(1)}_{v}=1,v\in V_{G_{S}};m^{(1)}_{v}=0,v\in V_{G}/V_{G_{S}}}
\\&=\sum_{v\in V_{G_{S}}}\mathbf h^{(1)}_{v}+\sum_{v\in  V_{G}/V_{G_{S}}}\mathbf h^{(1)}_{v}
\\&=\sum_{v\in V_{G_{S}}}\textrm{ReLU}(\mathbf W^{(1)}_{1}[\mathbf x_{v}||\sum_{u\in\mathcal{N}_{G_{S}}(v)}\mathbf x_{u}])+\mathbf 0
\\&=\mathbf h_{G_{S}}|_{m^{(1)}_{v}=1,v\in V_{G_{S}}}.
\end{aligned}
\end{equation}
It removes the restriction of general GNNs that the aggregation is conducted on all nodes, and therefore we call it as $sparse$ aggregation.
The selected subgraph includes all nodes with mask 1 and related edges.

In a multi-layer SMG, nodes with $m^{(k)}_{v}=0$ are not actually removed.
The GNN operation in the following layers would also involve them.
In order to represent the subgraph $G_{S}$, a basic idea is to restrict all layers to learn on the same subgraph $G_{S}$ by assigning $m^{(k)}_{v}=1$ for all $v\in V_{G_{S}}$ and $m^{(k)}_{u}=0$ for all $u\in V_{G}/V_{G_{S}}$ in all layers.
This corresponds to a specific subgraph sequence that all subgraphs are the same.
However, not all masks assigned on nodes in different layers have an effect on the final representations, and masks in some layers can take arbitrary values.
Since different masks lead to a different subgraph sequence, this means that the required subgraph representation may correspond to a different subgraph sequence.
To explain this,
we use $\mathbf H^{(k)}_{G}=\{\{\mathbf h^{(k)}_{v}|v\in V_{G}\}\}=\textrm{SMG}_{k}(\mathbf X_{G},\mathbf A_{G},\mathbf M)$ to denote the set of node representations learned by a $k$-layer SMG, where the input $\mathbf M\in[0,1]^{k\times n}$ is the preassigned masks at all $k$ layers with $\mathbf M_{k,v}=m^{(k)}_{v}$.
$\mathbf M$ defines a sequence of subgraphs of length $k$.
$\hat{\mathbf H}^{(k)}_{G}=\{\{\hat{\mathbf h}^{(k)}_{v}|v\in V_{G}\}\}=\textrm{SMG}_{k}(\mathbf X_{G},\mathbf A_{G},\mathbf 1)$ is the set of node representations learned by $k$-layer SMG with $\mathbf M=\mathbf 1^{k\times n}$,
and correspondingly $\hat{\mathbf h}^{(k)}_{v}=\textrm{ReLU}(\mathbf W^{(k)}_{1}[\hat{\mathbf h}^{(k-1)}_{v}||\sum_{u\in\mathcal N(v)}\hat{\mathbf h}^{(k-1)}_{u}])$.
\begin{lemma}
For any subgraph $G_S$ of $G$,
Let $\mathbf H^{(K)}_{G}=\{\{\mathbf h^{(k)}_{v}|v\in V_{G}\}\}$ and $\hat{\mathbf H}^{(K)}_{G_{S}}=\{\{\hat{\mathbf h}^{(k)}_{v}|v\in V_{G_{S}}\}\}$. Then we have $\mathbf h^{(K)}_{v}=\hat{\mathbf h}^{(K)}_{v}$ for any $v\in V_{G_{S}}$, if the following condition holds,
\begin{equation}
\begin{aligned}
\nonumber
\mathbf M_{k,v}=m^{(k)}_{v}=
\begin{cases}
0&\{v\}\subseteq V_{G}/V_{G_{S}}
  \\&\wedge \mathcal{N}_{G}(v)\cap V_{G_{S}}\neq\emptyset
  \\&\wedge k\%2=1\\
1&\{v\}\subseteq V_{G_{S}}\wedge \{k\}\subseteq[K].
\end{cases}
\end{aligned}
\end{equation}
\label{lemma1}
\end{lemma}
We prove Lemma \ref{lemma1} in Appendix \ref{proof-lemma1}.
Lemma \ref{lemma1} shows that we can obtain the node representations that are equivalent to node representations of any subgraph by controlling the assignments of $\mathbf M$.
Meanwhile, $\mathbf M$ (or the subgraph sequences) is not unique for the given subgraph.
This subgraph extraction mechanism takes effects implicitly by imposing restrictions on a combination of linear transformations, non-linear activation functions and aggregators.
Compared with existing graph sampling \cite{lee2018graph} or graph pooling \cite{gao2019graph,lee2019self,cangea2018towards,knyazev2019understanding},
the benefit is that it does not require setting a fixed sample or drop ratio and can extract the desired subgraph with arbitrary sizes.
For learning the graph-level representations, we have the following theorem.
\begin{theorem}
Let $\mathbf h_{G}=\textrm{SUM}(\mathbf H^{(K)}_{G})$ and $\hat{\mathbf h}_{G_{S}}=\textrm{SUM}(\hat{\mathbf H}^{(K)}_{G_{S}})$, where $G_{S}$ can be any subgraph of $G$. Then there exist assignments of $\mathbf M$ such that $\mathbf h_{G}=\hat{\mathbf h}_{G_{S}}$.
\label{theorem1}
\end{theorem}
We prove Theorem \ref{theorem1} in Appendix \ref{proof-theorem1}.
Lemma \ref{lemma1} and Theorem \ref{theorem1} indicate that the representation of any given subgraph corresponds to a group of $\mathbf M$ (or a group of subgraph sequences).
Then, the problem of representing the desired subgraphs is converted to finding proper $\mathbf M$.

\subsection{Hierarchical Representation Learning}
\label{hierarchical-section}

\begin{figure}[h]
\centering
\includegraphics[width=0.98\linewidth]{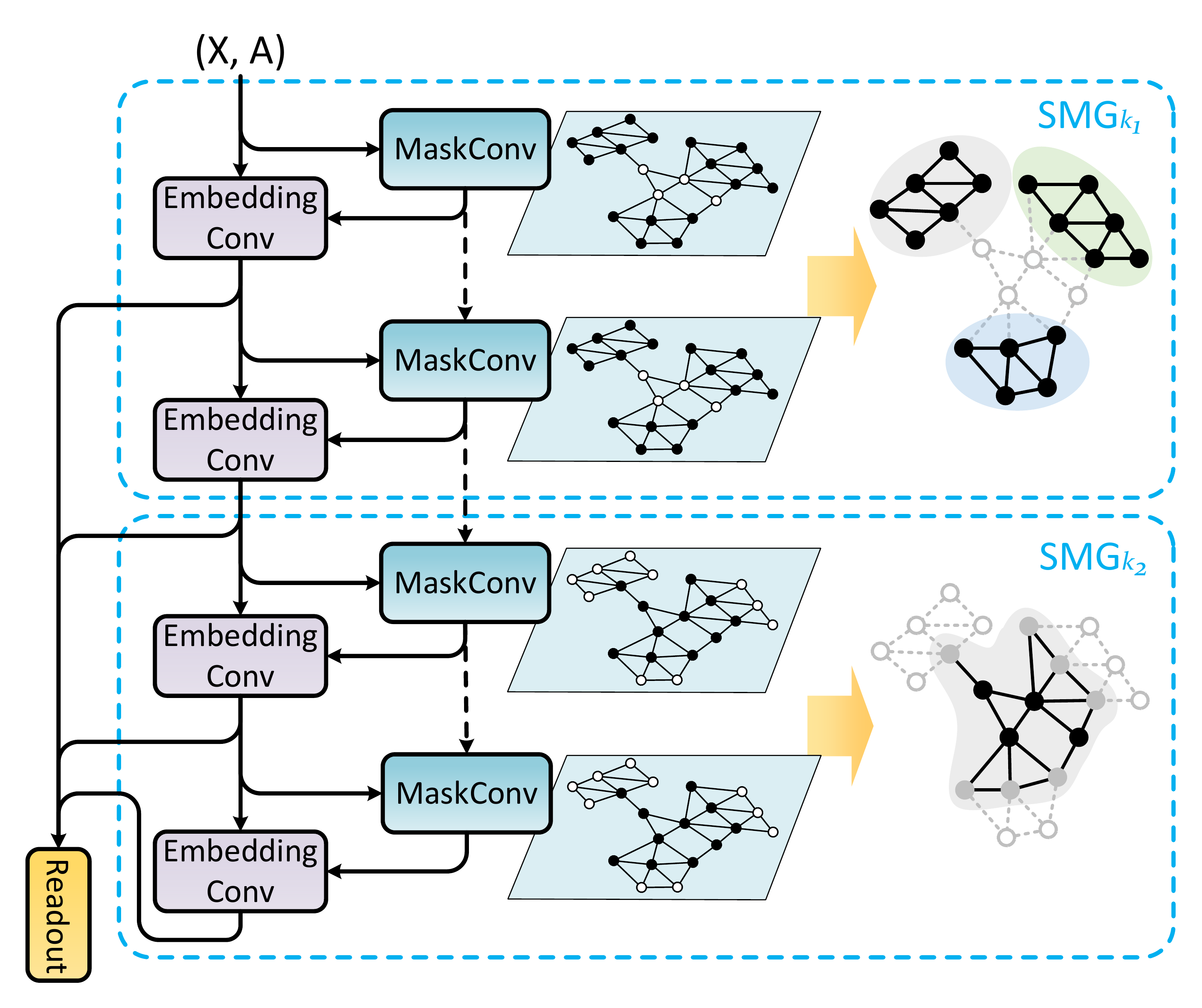}
\caption{Capture the hierarchical structures by stacking multiple SMGs.}
\label{Hierarchical:hierarchy}
\end{figure}
Graphs built from the real-world can inherently have hierarchical structures.
In the previous section, we have shown how a multi-layer SMG represents a given subgraph with proper assignments of $\mathbf M$.
In this section, we show how to control the assignments of $\mathbf M$ to capture the hierarchical structures of a graph.

A hierarchical graph as given in Figure \ref{Intro:hierarchy}(a) is characterized by each individual part and their interactions.
Meanwhile, each part is composed of interactions of several smaller parts, making the graph structure identification recursive.
To capture hierarchical structures, the neural network should first encode each local part individually
and then encodes their interactions.
From the node perspective, leaf nodes are aggregated first and then root nodes are taken into consideration.
This corresponds to stacking multiple SMGs,
where each $\textrm{SMG}_{k_{i}}$ learns node representations of the subgraph $G_{S_{i}}$ with input node features being node representations learned by previous $\textrm{GNN}_{k_{i-1}}$.
In the $i$-th $\textrm{GNN}_{k_{i}}$,
$\mathbf H^{\star}_{G}=\textrm{SMG}_{k_{i}}(\mathbf H^{\star-k_{i}}_{G},\mathbf A_{G},\mathbf M_{G_{S_{i}}})$, where $\mathbf M_{G_{S_{i}}}$ is the assignment corresponding to $G_{S_{i}}$ according to Lemma \ref{lemma1},
and $k_{i}$ is the number of layers of $\textrm{SMG}_{k_{i}}$. In the initial step, $\mathbf H^{0}_{G}=\mathbf X_{G}$.

We explain this process with the hierarchical graph example in Figure \ref{Intro:hierarchy}.
The hierarchical structure is characterized by three individual parts and their interactions.
We use a stacked 2 $\textrm{SMGs}$ to learn on this graph as showed in Figure \ref{Hierarchical:hierarchy}.
Based on the observation of Lemma \ref{lemma1},
the representation learned by $\textrm{SMG}_{k_{1}}$ corresponds to subgraph 1 with only leaf nodes involved, leading to the learned node representations limited within the corresponding part.
Also, the representation learned by $\textrm{SMG}_{k_{2}}$ corresponds to subgraph 2 consisting of root nodes (colored with black) and leaf nodes (colored with gray) which capture each part as well as their interactions.
Consistently, to extend to hierarchical graphs with heights larger than 2, stacking more $\textrm{SMGs}$ with each one learning on a specific subgraph is required.
Interactions of different parts at low levels are captured by lower SMGs, and those at high levels are captured by higher SMGs.

We have shown how to capture hierarchical structures by stacking multi $\textrm{SMGs}$ with each one corresponding to a subgraph as analyzed in Lemma \ref{lemma1}.
This assumes the knowledge of the hierarchical structures of the graph.
However, the hierarchical structures are not a priori knowledge and should be captured by the neural network itself.
Thanks to
\begin{equation}
\nonumber
\begin{aligned}
\mathbf H^{(K)}_{G}&=\textrm{SMG}_{k_{L}}(\mathbf H^{(K-k_{L})}_{G},\mathbf A_{G},\mathbf M_{G_{S_{L}}})
\\&=\textrm{SMG}_{k_{L}}(\textrm{SMG}_{_{k_{L-1}}}(...\textrm{SMG}_{k_{1}}(\mathbf X_{G},\mathbf A_{G},\mathbf M_{G_{S_{1}}}),
\\&\quad\ \mathbf A_{G},\mathbf M_{G_{S_{L-1}}}),\mathbf A_{G},\mathbf M_{G_{S_{L}}})
\\&=\textrm{SMG}_{K}(\mathbf X_{G},\mathbf A_{G},\mathop {||}^{L}_{i=1}\mathbf M_{G_{S_{i}}})
\\&=\textrm{SMG}_{K}(\mathbf X_{G},\mathbf A_{G},\mathbf M_G),
\end{aligned}
\label{T}
\end{equation}
which means that any required stacked $L$ SMGs with each one learning on a specific subgraph are equivalent to the same number layers of SMG with mask assignments $\mathbf M_G=\mathop {||}^{L}_{i=1}\mathbf M_{G_{S_{i}}}$.
We can obtain the required stacked $L$ SMGs with proper assignments of $\mathbf M_G$.
Then, the problem of capturing hierarchical structures is converted to finding the assignments of $\mathbf M_G$.

The flexibility of $\mathbf M_G$ enables SMG to extract much richer structural information, beyond the general dense aggregation based GNNs.
We have shown that a single layer SMG can only capture subgraphs, while a multi-layer SMG can capture hierarchical structures of graphs.
When $L>1$ and $G_{S_{i}}$ is restricted to be a subgraph of $G_{S_{i-1}}$ for all $i\in[L-1]$, SMG works in a similar way as top-$k$ based graph poolings \cite{gao2019graph,lee2019self,cangea2018towards,knyazev2019understanding} that iteratively remove some nodes and related edges layer by layer. 
However, the interactions of different parts as given in Figure \ref{Intro:hierarchy} cannot be captured by this kind of scheme,
since previously skipped nodes in lower layers will not be involved in higher layers.

\subsection{Mask Assignments Computations}
\label{soft-mask}

In this section, we present the method to compute the assignments of $\mathbf M_G$.
We use a separate GNN layer, called GNN-$w$ to learn the assignments of $\mathbf M_G$.
The value assigned on node $v$ in the $k$-th layer is
\begin{equation}
\begin{aligned}
\mathbf {M_{G}}_{k,v}&=m^{(k)}_{v}
\\&=\textrm{MLP}^{(k)}(\textrm{ReLU}([\mathbf L^{(k)}_{1}(m^{(k-1)}_{v}\mathbf h^{(k-1)}_{v})||
\\&\ \ \ \ \sum_{u\in\mathcal N(v)}\mathbf L^{(k)}_{2}(m^{(k-1)}_{u}\mathbf h^{(k-1)}_{u})])),
\end{aligned}
\label{score}
\end{equation}
where $\mathbf L^{(k)}_{1}$ and $\mathbf L^{(k)}_{2}$ are affine maps, MLP is a $2$-layer perceptron with output feature dimension of 1.
The activation function in the last layer of MLP is Sigmoid, making $m^{(k)}_{v}\in(0,1)$.
Note that the constant parts (i.e. bias) in $\mathbf L^{(k)}_{1}$, $\mathbf L^{(k)}_{2}$ and MLP is required.
Otherwise, $m^{(k-1)}_{v}\mathbf h^{(k-1)}_{v}=\mathbf0$ and $m^{(k-1)}_{u}\mathbf h^{(k-1)}_{u}=\mathbf0$ will lead $m^{(k)}_{v}$ always fixed to 0.5.

The value of $m^{(k)}_{v}$ assigned on each node is from 0 to 1, not discrete mask value 0 or 1 as expected.
Fortunately, in a SMG layer, $\mathbf h^{(k)}_{v}=f(\{m^{(k)}_{i}|i\in\hat{\mathcal{N}}(v)\})=\textrm{ReLU}(\mathbf W^{(k)}_{1}m^{(k)}_{v}[\mathbf h^{(k-1)}_{v}||\\\sum_{u\in\mathcal N(v)}m^{(k)}_{u}\mathbf h^{(k-1)}_{u}])$ is continuous for $m^{(k)}_{i}\in [0,1]$ where $i\in\hat{\mathcal{N}}(v)$.
Therefore we have $\lim\limits_{m^{(k)}_{i}\to 0^{+}}f(m^{(k)}_{i},...)=f(0,...)$ and\\$\lim\limits_{m^{(k)}_{i}\to 1^{-}}f(m^{(k)}_{i},...)=f(1,...)$ for any $i\in\hat{\mathcal{N}}(v)$, making it possible to use $m^{(k)}_{v}$ to approximate the mask value 0 or 1.

The benefit of soft-mask is that the weights are taken into consideration.
As $\mathbf h^{(k)}_{v}$ represents a $k$-hop subtree rooted on $v$ that captures the structural information of local $k$-hop substructure rooted on $v$, $m^{(k)}_{v}$ multiplies with $\mathbf h^{(k)}_{v}$ gives the weight of that substructure for the following aggregation operation that takes $\mathbf h^{(k)}_{v}$ as inputs.
For nodes with 0 weights, it is equivalent to drop them and related edges.

To maintain zero invariant, we use SUM as Readout function as presented in Theorem \ref{theorem1}.
We also use jumping concatenation as given in \cite{xu2018representation},
\begin{equation}
\begin{aligned}
\mathbf h_{G}=\mathop {||}^{K}_{k=1}(\sum^{N}_{i=1}\mathbf h^{(k)}_{i}).
\end{aligned}
\label{readout2}
\end{equation}
Equation \ref{readout2} utilizes node representations from different layers to compute the entire graph representation $\mathbf h_{G}$.
Since a node representation in the $k$-th layer $\mathbf h^{(k)}_{v}$ encodes the $k$-hop substructure with the weight $m^{(k)}_{v}$,
when implementing Readout function as Equation \ref{readout2},
the weight of $\mathbf h^{(k)}_{v}$ is considered over all substructures with different hops.

\subsection{Multi-channel Soft-mask GNN Model}

We generalize the soft-mask mechanism to a multi-channel scenario. 
To this end, GNN-$w$ computes a mask value for each channel of a node representation such that $m^{(k)}_{v}\in(0,1)^{d}$, where $d$ is the dimension of node representations.
This is done by setting the output feature dimension of MLP in Equation \ref{score} to be $d$.
Correspondingly, the SMG layer (Equation \ref{mask-gnn}) is rewritten as
\begin{equation}
\begin{aligned}
\mathbf h^{(k)}_{v}=&\textrm{ReLU}(m^{(k)}_{v}\odot \mathbf W^{(k)}_{2}[\mathbf h^{(k-1)}_{v}||\\&\sum_{u\in\mathcal N(v)}m^{(k)}_{u}\odot \mathbf h^{(k-1)}_{u}]),
\end{aligned}
\label{mc-mask-gnn}
\end{equation}
where $\odot$ is  element-wise multiplication, and $\mathbf W^{(k)}_{2}\in\mathbf R^{d\times d}$.
Before the GNN operation in the first layer, we use a linear transformation to make the dimension of the input node features equal to the number of hidden units.
Intuitively, multi-channel SMG allows us to conduct sparse aggregation on each channel respectively.
General SMG can be regarded as a special case of multi-channel SMG where the values on each channel are the same.

\section{Discussion}

\textbf{Comparisons with GAT.}
In GAT, the attention coefficients of a node are different for its neighbors.
This is because, in node level predictions, a node has varying impacts to its neighbors.
For graph level predictions, the aim is to embed node features as well as graph topology,
and therefore the impact of nodes should be considered over the entire graph.
We can see that GAT applies $neighborhood$ attention for learning node representations, while SMG applies $global$ attention for learning the entire graph representation.
Also note that due to the neighborhood normalization in GAT, GAT cannot completely skip some nodes and related edges.
In our experiments, we give detailed comparisons about these two kinds of strategies on graph level tasks, even though there are relatively few attempts that apply GAT to graph level predictions.

\textbf{Comparisons with top-$k$ based poolings.} 
Top-$k$ based poolings \cite{gao2019graph,lee2019self,cangea2018towards,knyazev2019understanding} compute the attention coefficients among all nodes, making the learned representations focus on informative parts.
To handle this, all top-$k$ based methods physically remove nodes and related edges with low attention weights.
The drawback of this kind of mechanism is that it requires to manually set the drop-ratio, which can be a limitation for capturing desired subgraphs with arbitrary sizes.
Soft-mask mechanism is proposed without this limitation, and therefore SMG should be more adaptive in graphs with arbitrary scales.
Furthermore, in top-$k$ based poolings, the learned subgraph of the current layer is the subgraph of the previous layer.
This can be a limitation for the model to extract hierarchical structures, as we explained in Section \ref{hierarchical-section}.

\section{Experiments}

In this section, we evaluate soft-mask GNN and its variants on both graph classification and graph regression tasks.
The code is available at \url{https://github.com/qslim/soft-mask-gnn}.

\subsection{Datasets}

For graph classification task, our experiments are conducted on 10 real-world graph datasets from \cite{yanardag2015deep,KKMMN2016}, including both bioinformatics datasets and social network datasets. Detailed statistics are given in Table \ref{classification-results}.
For graph regression task, we use QM9 dataset \cite{ramakrishnan2014quantum,wu2018moleculenet,ruddigkeit2012enumeration}, which is composed of 134K small organic molecules. The task is to predict 12 targets for each molecule.
All data is obtained from pytorch-geometric library \cite{Fey/Lenssen/2019}.

\subsection{Graph Classification Task}

\begin{table*}[t]
\caption{Graph classification results. The top 2 performance approaches are highlighted in bold. We report the results in the original papers by default. When the results are not given in the original papers, we report the best testing results given in \protect\cite{zhang2018end,pmlr-v80-ivanov18a,xinyi2018capsule}.}
\label{classification-results}
\centering
\resizebox{0.98\textwidth}{!}{
\scriptsize
\begin{tabular}{lllllllllll}
\toprule
dataset        & MUTAG                           & PROTEINS                        & NCI1                            & COLLAB                          & ENZYMES                         & IMDB-B                          & IMDB-M                          & RDT-B                           & RDT-M5K                         & RDT-12K                         \\ \midrule
\# graphs      & 188                             & 1113                            & 4110                            & 5000                            & 600                             & 1000                            & 1500                            & 2000                            & 4999                            & 11929                           \\
\# classes     & 2                               & 2                               & 2                               & 3                               & 6                               & 2                               & 3                               & 2                               & 5                               & 11                              \\
avg \# nodes   & 17.9                            & 39.1                            & 29.9                            & 74.5                            & 32.6                            & 19.8                            & 13.0                            & 429.6                           & 508.5                           & 391.4                           \\ \midrule
GK \cite{shervashidze2009efficient}             & 81.58$\pm$2.11                  & 71.67$\pm$0.55                  & 62.49$\pm$0.27                  & 72.84$\pm$0.28                  & 32.70$\pm$1.20                  & 65.87$\pm$0.98                  & 43.89$\pm$0.38                  & 77.34$\pm$0.18                  & 41.01$\pm$0.17                  & 31.82$\pm$0.08                  \\
RW \cite{vishwanathan2010graph}                 & 79.17$\pm$2.1                   & 59.57$\pm$0.1                   & NA                              & NA                              & 24.16$\pm$1.64                  & NA                              & NA                              & NA                              & NA                              & NA                              \\
PK \cite{neumann2016propagation}                & 76$\pm$2.7                      & 73.68$\pm$0.7                   & 82.54$\pm$0.5                   & NA                              & NA                              & NA                              & NA                              & NA                              & NA                              & NA                              \\
WL \cite{shervashidze2011weisfeiler}            & 84.11$\pm$1.9                   & 74.68$\pm$0.5                   & $\textbf{84.46}\pm\textbf{0.5}$ & NA                              & 52.22$\pm$1.26                  & 73.40$\pm$4.63                  & 49.33$\pm$4.75                  & 81.0$\pm$3.10                   & 49.44$\pm$2.36                  & 38.18$\pm$1.30                  \\
FGSD \cite{verma2017hunt}                       & $\textbf{92.12}$                & 73.42                           & 79.80                           & 80.02                           & NA                              & 73.62                           & $\textbf{52.41}$                & NA                              & NA                              & NA                              \\
AWE \cite{pmlr-v80-ivanov18a}                  & 87.87$\pm$9.76                  & NA                              & NA                              & 73.93$\pm$1.94                  & 35.77$\pm$5.93                  & 74.45$\pm$5.83                  & 51.54$\pm$3.61                  & 87.89$\pm$2.53                  & 50.46$\pm$1.91                  & 39.20$\pm$2.09                  \\
DGCNN \cite{zhang2018end}                       & 85.83$\pm$1.66                  & 75.54$\pm$0.94                  & 74.44$\pm$0.47                  & 73.76$\pm$0.49                  & 51.0$\pm$7.29                   & 70.03$\pm$0.86                  & 47.83$\pm$0.85                  & NA                              & NA                              & NA                              \\
PSCN \cite{niepert2016learning}                 & 88.95$\pm$4.4                   & 75$\pm$2.5                      & 74.44$\pm$0.5                   & 73.76$\pm$0.5                   & NA                              & 45.23$\pm$2.84                  & 45.23$\pm$2.84                  & 86.30$\pm$1.58                  & 49.10$\pm$0.70                  & 41.32$\pm$0.42                  \\ \midrule
DCNN \cite{atwood2016diffusion}                 & NA                              & 61.29$\pm$1.60                  & 56.61$\pm$1.04                  & 52.11$\pm$0.71                  & NA                              & 49.06$\pm$1.37                  & 33.49$\pm$1.4                   & NA                              & NA                              & NA                              \\
ECC \cite{simonovsky2017dynamic}                & 76.11                           & NA                              & 76.82                           & NA                              & 45.67                           & NA                              & NA                              & NA                              & NA                              & NA                              \\
DGK \cite{yanardag2015deep}                     & 87.44$\pm$2.72                  & 75.68$\pm$0.54                  & 80.31$\pm$0.46                  & 73.09$\pm$0.25                  & 53.43$\pm$0.91                  & 66.96$\pm$0.96                  & 44.55$\pm$0.52                  & 78.04$\pm$0.39                  & 41.27$\pm$0.18                  & 32.22$\pm$0.10                  \\
CapsGNN \cite{xinyi2018capsule}                 & 86.67$\pm$6.88                  & 76.28$\pm$3.63                  & 78.35$\pm$1.55                  & 79.62$\pm$0.91                  & 54.67$\pm$5.67                  & 73.10$\pm$4.83                  & 50.27$\pm$2.54                  & NA                              & 52.88$\pm$1.48                  & 46.62$\pm$1.90                  \\
DiffPool \cite{ying2018hierarchical}            & NA                              & 76.25                           & NA                              & 75.48                           & $\textbf{62.53}$                & NA                              & NA                              & NA                              & NA                              & 47.08                           \\
GIN \cite{xu2018how}                       & 89.4$\pm$5.6                    & 76.2$\pm$2.8                    & 82.7$\pm$1.7                    & 80.2$\pm$1.9                    & NA                              & $\textbf{75.1}\pm\textbf{5.1}$  & 52.3$\pm$2.8                    & 92.4$\pm$2.5                    & $\textbf{57.5}\pm\textbf{1.5}$  & NA                              \\
Top-$k$ Pool \cite{lee2019self}                      & NA                              & 71.86$\pm$0.97                  & 67.45$\pm$1.11                  & NA                              & NA                              & NA                              & NA                              & NA                              & NA                              & NA                              \\
PPGNN \cite{maron2019provably}                  & $\textbf{90.55}\pm\textbf{8.7}$ & $\textbf{77.2}\pm\textbf{4.73}$ & $83.19\pm1.11$                  & 80.16$\pm$1.11                  & NA                              & 72.6$\pm$4.9                    & 50$\pm$3.15                     & NA                              & NA                              & NA                              \\ \midrule
SMG            & 89.2$\pm$6.22                   & $\textbf{76.8}\pm\textbf{4.15}$ & $\textbf{83.3}\pm\textbf{1.88}$ & $\textbf{83.2}\pm\textbf{1.60}$ & 59.0$\pm$5.07                   & 74.8$\pm$6.63                   & 52.0$\pm$4.11                   & $\textbf{92.9}\pm\textbf{2.84}$ & 57.3$\pm$2.09                   & $\textbf{51.2}\pm\textbf{1.40}$ \\
SMG-JK         & 89.3$\pm$7.12                   & 76.3$\pm$4.57                   & 83.0$\pm$2.25                   & $\textbf{82.7}\pm\textbf{1.57}$ & $\textbf{60.3}\pm\textbf{5.26}$ & 75.0$\pm$6.24                   & 52.3$\pm$3.20                   & $\textbf{93.1}\pm\textbf{2.25}$ & $\textbf{57.5}\pm\textbf{1.56}$ & $\textbf{51.3}\pm\textbf{1.52}$ \\
M-SMG-JK       & 89.6$\pm$7.38                   & 76.1$\pm$4.04                   & 82.8$\pm$1.73                   & 82.6$\pm$1.54                   & 57.3$\pm$5.63                   & $\textbf{75.0}\pm\textbf{5.95}$ & $\textbf{52.7}\pm\textbf{3.71}$ & 91.7$\pm$2.08$^{*}$             & 57.0$\pm$1.85$^{*}$             & 49.2$\pm$1.13$^{*}$             \\
Rank           & $3^{rd}$                        & $2^{nd}$                        & $2^{nd}$                        & $1^{st}$                        & $2^{nd}$                        & $2^{nd}$                        & $1^{st}$                        & $1^{st}$                        & $1^{st}$                        & $1^{st}$                        \\ \bottomrule
\end{tabular}
}
\end{table*}

We follow the standard ways to evaluate models on classification datasets which perform 10-fold cross validation and report average accuracy \cite{xu2018how,ying2018hierarchical,maron2019provably}.
We use Adam optimizer \cite{kingma2014adam} with learning rate $\in\{0.005,0.001,0.0005\}$ and learning rate decay $\in [0.7, 1]$ every $\{50,100\}$ epochs.
Other hyperparameters settings are:
(1) the number of hidden units $\in\{32,64,128\}$ for bioinformatics datasets, and $\{64,128,256\}$ for social network datasets;
(2) the number of layers $\in\{2,3,4,5\}$;
(3) the batch size $\in\{64,128\}$;
(4) the dropout ratio $\in\{0,0.5\}$.
We run 10 independent times with selected hyperparameters to obtain the final accuracy on each dataset.
For graphs without node attributes, we use one-hot encoding of node degrees.

We evaluate the following variants of our proposed soft-mask GNN.
\textbf{SMG} represents soft-mask GNN with GNN-$w$ implemented by Equation \ref{score}.
\textbf{M-SMG} represents multi-channel soft-mask GNN implemented by Equation \ref{mc-mask-gnn}.
\textbf{JK} represents the Readout operation implemented by Equation \ref{readout2}.

\textbf{Comparisons with baselines.}
Table \ref{classification-results} presents a summary of classification results.
Baselines include kernel-based methods and GNN methods.
The last row indicates the ranking of our models.
M-SMG-JK fails to converge on REDDIT due to a large number of nodes, and therefore their results (marked with $^*$) are computed with the MEAN readout function.
Note that SMG based models achieve the top 2 performance in 9 out of 10 datasets.
Especially, SMG and SMG-JK are more competitive on large graphs, e.g., COLLAB and three REDDIT datasets.
This is because, in a large graph, task-relevant structures should not be characterized in every single detailed structure.
It is more efficient to focus on the relevant substructures or the hierarchical structure of a graph.

\textbf{Representation Power of SMG Models.}
\begin{figure*}[h]
\centering
\includegraphics[width=0.85\textwidth]{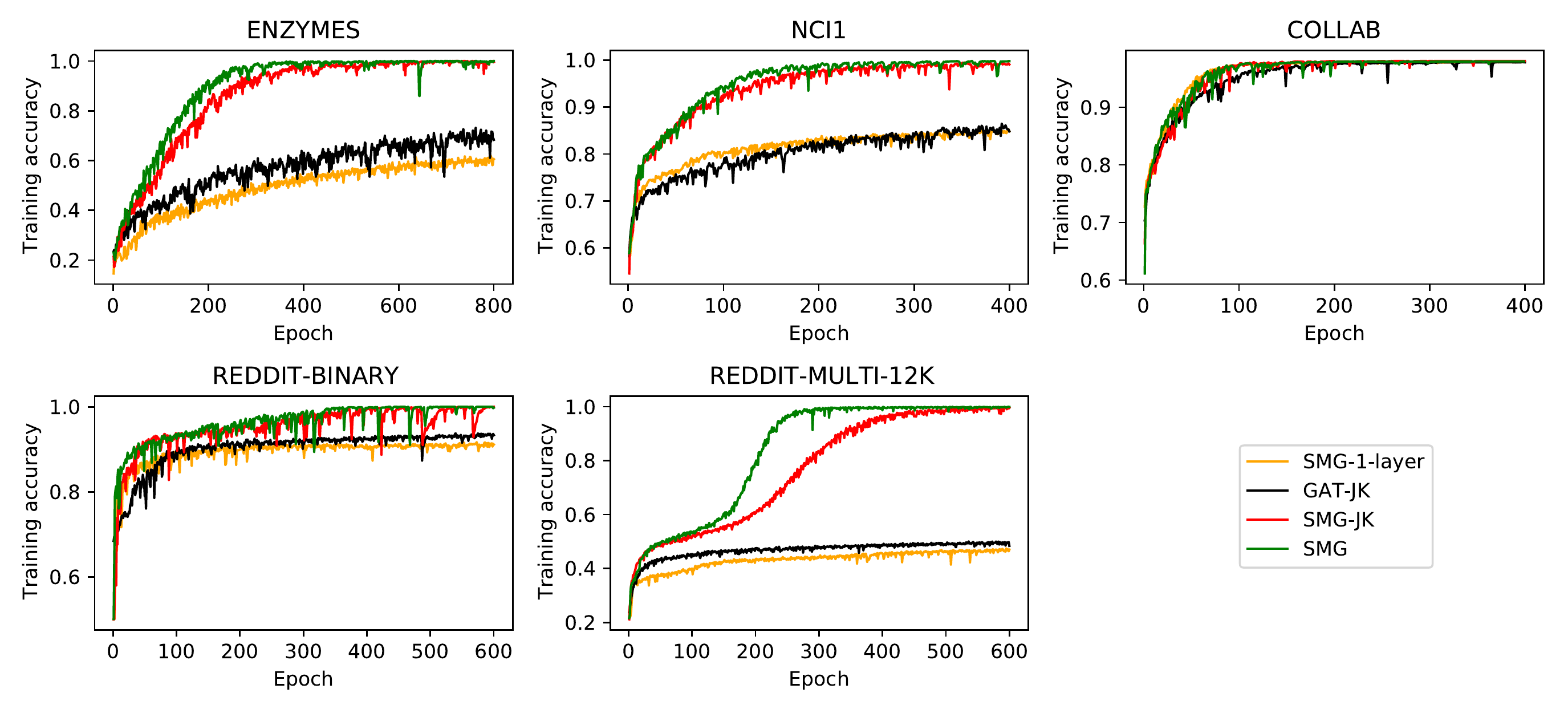}
\caption{Training set performance.}
\label{training-accuracy}
\end{figure*}
Figure \ref{training-accuracy} gives training set performance.
Since our proposed soft mask mechanism tries to minimize the impact of irrelevant parts and only learns from the desired subgraph in each layer, we can see that SMG and SMG-JK boost training set performance significantly. 
Our models fit especially well on large and sparse graphs as given in REDDIT datasets, where GAT fails to fit them.
The multi-layer SMG always outperforms SMG-1-layer, except for COLLAB where both models achieve the highest accuracies.
This is because graphs in COLLAB have dense connections as visualized in Figure \ref{visual-COLLAB} in Appendix \ref{visualizing},
and exploring 1-hop neighbors can be sufficient to get an overview of graphs.
For large and sparse graphs, structural information may be characterized by long-range dependencies that can be captured by exploring larger hops in higher layers.
GAT suffers from the impact of irrelevant parts, making them not benefit from the deeper structures used for capturing long-range dependencies.
The performance differences between GAT and SMG show that skipping noisy nodes and related edges makes the model better fit training data.

\textbf{Visualizing Weight Distributions.}
\begin{figure*}[h]
\centering
\includegraphics[width=0.5\textwidth]{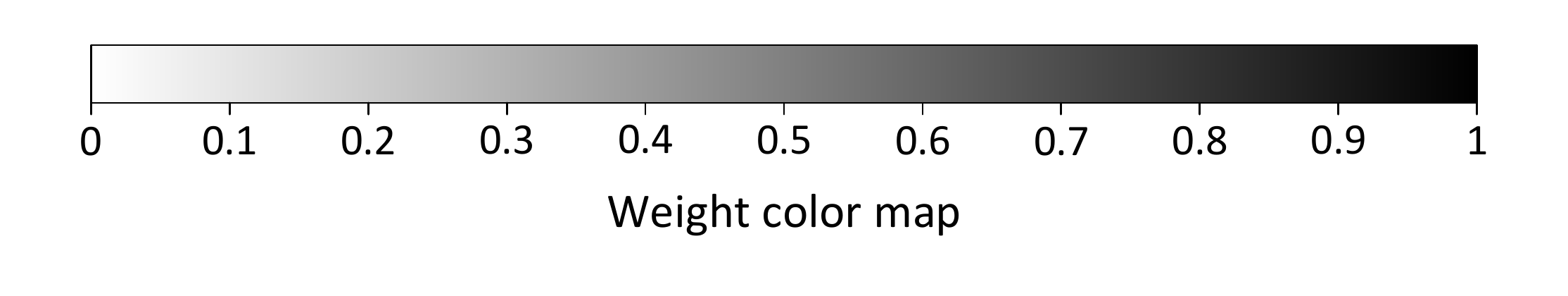}
\includegraphics[width=0.82\textwidth]{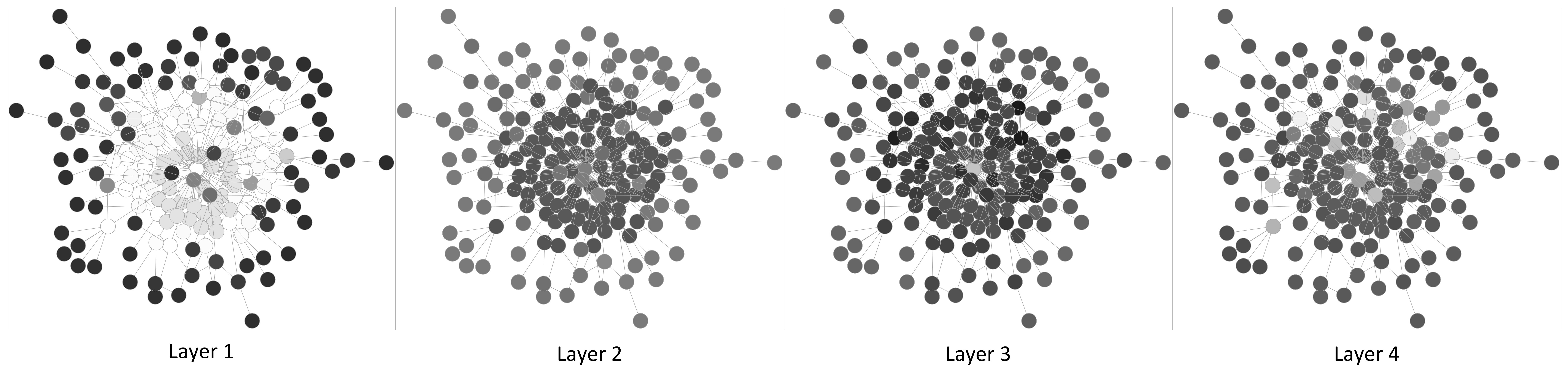}
\includegraphics[width=0.82\textwidth]{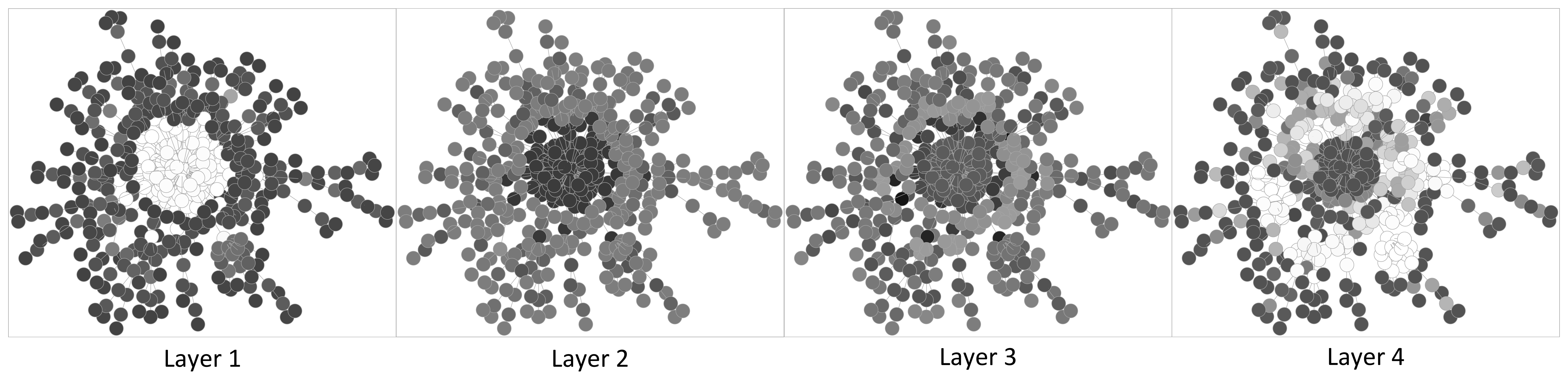}
\includegraphics[width=0.82\textwidth]{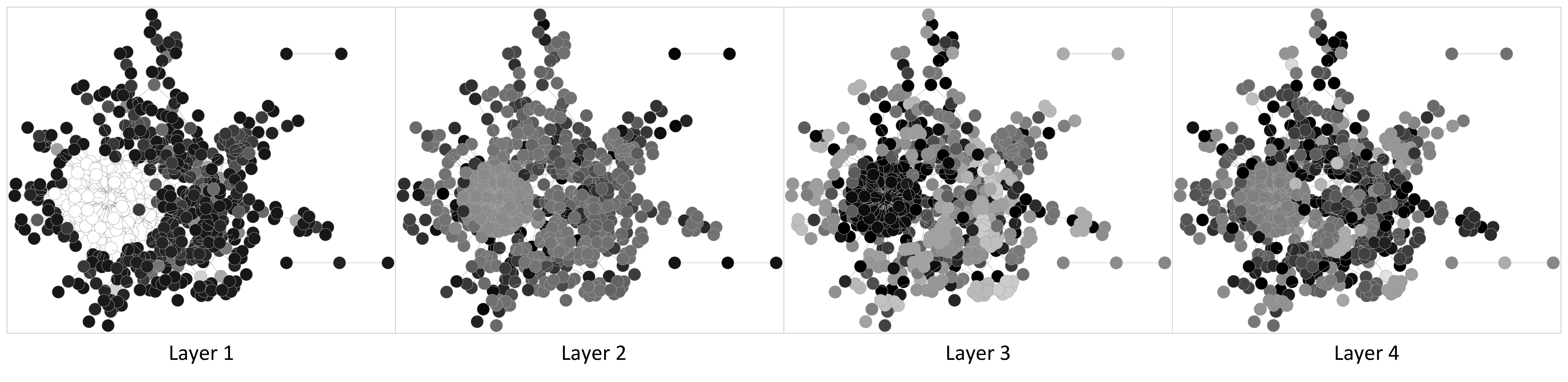}
\includegraphics[width=0.82\textwidth]{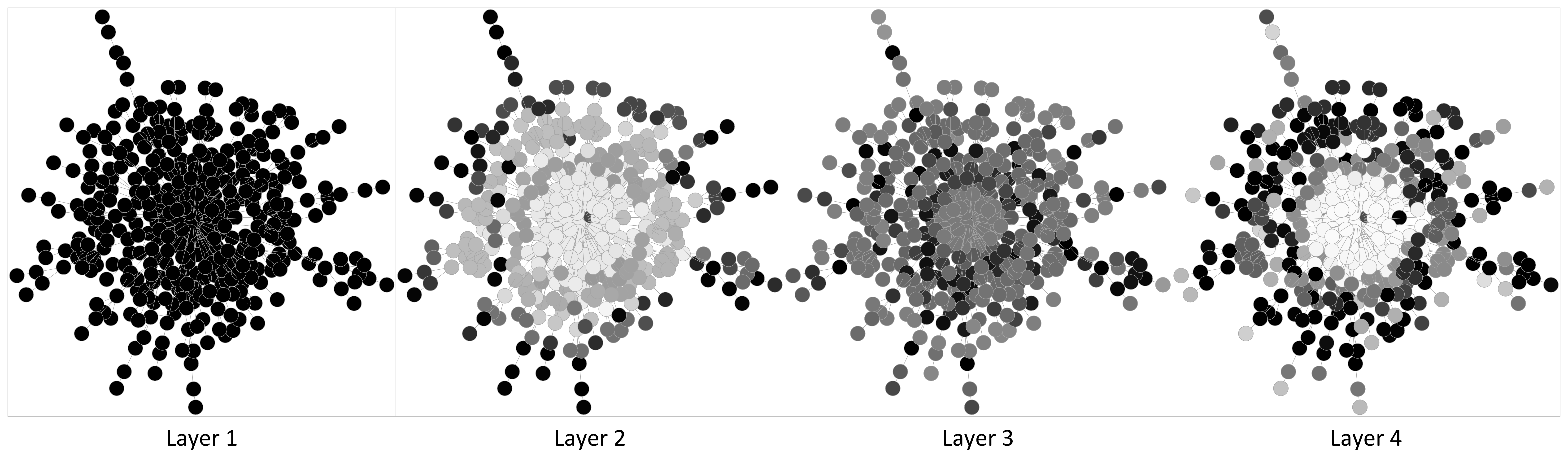}
\caption{Weight distributions of 4 graphs sampled from REDDIT-BINARY, REDDIT-MULTI-5K and REDDIT-MULTI-12K.}
\label{visual-RDT}
\end{figure*}
\begin{table*}[h]
\caption{Mean absolute errors on the QM9 dataset. The top 2 best-performing GNNs are highlighted in bold.}
\label{regression-results}
\centering
\resizebox{0.85\textwidth}{!}{
\scriptsize
\begin{tabular}{lllllllllllll}
\toprule
target   & $\mu$             & $\alpha$          & $\epsilon_{homo}$  & $\epsilon_{lumo}$ & $\Delta \epsilon$ & $\langle R^2 \rangle$ & $ZPVE$              & $U_0$             & $U$               & $H$               & $G$               & $C_{v}$           \\
\midrule
DTNN \cite{wu2018moleculenet}    & $\textbf{0.244}$  & 0.95              & 0.00388            & 0.00512           & 0.0112            & $\textbf{17}$         & 0.00172             & 2.43              & 2.43              & 2.43              & 2.43              & 0.27              \\
MPNN \cite{gilmer2017neural}     & 0.358             & 0.89              & 0.00541            & 0.00623           & 0.0066            & 28.5                  & 0.00216             & 2.05              & 2                 & 2.02              & 2.02              & 0.42              \\
PPGNN \cite{maron2019provably}   & $\textbf{0.0934}$ & $\textbf{0.318}$  & $\textbf{0.00174}$ & $\textbf{0.0021}$ & $\textbf{0.0029}$ & $\textbf{3.78}$       & 0.000399            & $\textbf{0.022}$  & 0.0504            & 0.0294            & 0.024             & 0.144             \\
\midrule
SMG-JK   & 0.4709            & 0.3415            & 0.0033             & 0.0036            & 0.0049            & 23.64                 & $\textbf{0.000236}$ & 0.0248            & $\textbf{0.0247}$ & $\textbf{0.0225}$ & 0.0244            & $\textbf{0.130}$  \\
M-SMG-JK & 0.4395            & $\textbf{0.2899}$ & $\textbf{0.0030}$  & $\textbf{0.0032}$ & $\textbf{0.0045}$ & 21.90                 & $\textbf{0.000196}$ & $\textbf{0.0212}$ & $\textbf{0.0202}$ & $\textbf{0.0214}$ & $\textbf{0.0212}$ & $\textbf{0.1157}$ \\
Rank     & $4^{th}$          & $1^{st}$          & $2^{nd}$           & $2^{nd}$          & $2^{nd}$          & $3^{rd}$              & $1^{st}$            & $1^{st}$          & $1^{st}$          & $1^{st}$          & $1^{st}$          & $1^{st}$          \\
\bottomrule
\end{tabular}
}
\end{table*}
We give weights distributions
\footnote{Graphs are visualized by graphviz.
There are different layout engines provided by Graphviz.
For large graphs in REDDIT and COLLAB, we use sfdp to provide a friendly view.
For small graphs in PROTEINS and NCI1, we use neato to give detailed structure information.}
of REDDIT on all layers computed by SMG as shown in Figure 4.
More visualization results on other datasets are
provided in the Appendix.
We interpret weight distributions from two perspectives:
(1) How the weights are related to graph structures in a single layer;
(2) How weights change over different layers.
All graphs are randomly sampled from the corresponding datasets.
We use the depth of color to represent the weights of nodes.

From visualizations, the weights of nodes have significant differences and weight distributions do have strong relations to graph structures.
Graphs in REDDIT are characterized by tree-like structures, where most nodes lie in the root and are densely connected.
This is also reflected from the layout in Figure \ref{visual-RDT} where nodes close to the root are at the center of the layout and their dense connections form a clique with overlaps.
Viewed on different layers, the weights move from leaf parts to the root part.
In the first layer, the weights are completely distributed on leaf parts with almost zero weights on the root part, indicating that the learned node representations skip the root part.
In higher layers, the weights of the root part are increased. This can be the evidence that the hierarchical structures are captured by SMG.
Node representations in the last layer (Layer 4 in Figure \ref{visual-RDT}) are used to compute the final graph representation and the zero weights are not involved in the final graph representation since the Readout operation is zero invariant.
Weights distributions in Layer 4 show that only some of the node representations participate in the computation of the final graph representation.

Note that the weight differences become less significant in higher layers.
This is because the weight of a node actually characterizes the weight of a subtree rooted on that node.
In higher layers, the subtrees become larger and have more overlaps, making them similar.
On most datasets we test, weight distributions change dynamically in different layers.
The substructures assigned with low weights in lower layers can be assigned with high weights in higher layers, which is consistent with our analysis.
Note that this kind of phenomenon cannot be captured by top-$k$ based poolings.

\subsection{Graph Regression Task}
Following the standard dataset splits described in \cite{wu2018moleculenet,maron2019provably}, the QM9 dataset is randomly split into 80\% train, 10\% validation and 10\% test.
We evaluate SMG-JK and M-SMG-JK with the following parameters: number of layers $\in\{3,4\}$; hidden units = 64; batch size = 64; learning rate $\in\{0.001,0.0005,0.0001\}$; learning rate decay $\in[0.75, 1]$ every $\{20,30,50\}$ epochs. Both models are trained for 500 epochs.
Table \ref{regression-results} compares the mean absolute error of our methods with state-of-the-art approaches.
All results of these approaches are taken from the original papers.
Note that our methods achieve the lowest error on 7 out of the 12 targets.

\section{Conclusion}
Motivated by effectively skipping irrelevant parts of graphs,
we propose soft-mask GNN (SMG) layer, which learns graph representations from a sequence of subgraphs.
We show its capability for explicitly extracting desired substructures or hierarchical structures.
Experimental results on benchmark graph classification and graph regression datasets demonstrate that SMG gains significant improvements and the visualizations of masks provide interpretability of structures learned by the model.

\begin{acks}
This work is supported in part by the National Natural Science Foundation of China under Grants U1811463, 62072069, and U19B2039, and also in part by the Innovation Foundation of Science and Technology of Dalian under Grants 2018J11CY010 and 2019J12GX037.
\end{acks}

\bibliographystyle{ACM-Reference-Format}
\bibliography{main}

\appendix

\section{Proof of Lemma \ref{lemma1}}
\label{proof-lemma1}

\begin{proof}
When $K$=1, for any $v\in V_{G_{S}}$, $m^{(1)}_{v}=1$ and
\begin{equation}
\begin{aligned}
\nonumber
\mathbf h^{(1)}_{v}=&\textrm{ReLU}(\mathbf W^{(1)}_{1}\mathbf x_{v}+\sum_{u\in\mathcal N_{G}(v)\cap V_{G_{S}}}m^{(1)}_{u}\mathbf W^{(1)}_{2}\mathbf x_{u}+
\\&\sum_{u\in\mathcal N_{G}(v)\cap (V_{G}/V_{G_{S}})}m^{(1)}_{u}\mathbf W^{(1)}_{2}\mathbf x_{u}). 
\end{aligned}
\end{equation}
To prove $\sum_{u\in\mathcal N_{G}(v)\cap (V_{G}/V_{G_{S}})}m^{(1)}_{u}\mathbf W^{(1)}_{2}\mathbf x_{u}=\mathbf 0$,
we only need to consider the case that $\mathcal N_{G}(v)\cap (V_{G}/V_{G_{S}})\neq\emptyset$.
For any $u\in\mathcal N_{G}(v)\cap (V_{G}/V_{G_{S}})$, we have
(i) $\{u\}\subseteq V_{G}/V_{G_{S}}$;
(ii)$\mathcal{N}_{G}(u)\cap V_{G_{S}}\neq\emptyset$, since $\{v\}\subseteq\mathcal N_{G}(u)\cap V_{G_{S}}$;
(iii) $K\%2=1$.
According to given condition of $\mathbf M$, $m^{(1)}_{u}=0$. Then we have
\begin{equation}
\begin{aligned}
\nonumber
\mathbf h^{(1)}_{v}&=\textrm{ReLU}(\mathbf W^{(1)}_{1}\mathbf x_{v}+\sum_{u\in\mathcal{N}_{G}(v)\cap V_{G_{S}}}\mathbf W^{(1)}_{2}\mathbf x_{u}+\mathbf 0)
\\&=\textrm{ReLU}(\mathbf W^{(1)}_{1}\mathbf x_{v}+\sum_{u\in\mathcal{N}_{G_{S}}(v)}\mathbf W^{(1)}_{2}\mathbf x_{u})\\&=\mathbf t^{(1)}_{v}.
\end{aligned}
\end{equation}

Suppose when $K=k-1$, for any $v\in V_{G_{S}}$, $\mathbf h^{(k-1)}_{v}=\mathbf t^{(k-1)}_{v}$ holds. For any $v\in V_{G_{S}}$,
\begin{equation}
\begin{aligned}
\nonumber
\mathbf h^{(k)}_{v}=&\textrm{ReLU}(\mathbf W^{(k)}_{1}\mathbf h^{(k-1)}_{v}+\sum_{u\in\mathcal{N}_{G}(v)\cap V_{G_{S}}}m^{(k)}_{u}\mathbf W^{(k)}_{2}\mathbf h^{(k-1)}_{u}
\\&+\sum_{u\in\mathcal{N}_{G}(v)\cap (V_{G}/V_{G_{S}})}m^{(k)}_{u}\mathbf W^{(k)}_{2}\mathbf h^{(k-1)}_{u}).
\end{aligned}
\end{equation}
If $k\%2=1$, $m^{(k)}_{u}=0$ for any ${u\in\mathcal N_{G}(v)\cap (V_{G}/V_{G_{S}})}$ (the same as $K$=1) and $\sum_{u\in\mathcal{N}_{G}(v)\cap (V_{G}/V_{G_{S}})}m^{(k)}_{u}\mathbf W^{(k)}_{2}\mathbf h^{(k-1)}_{u}=\mathbf0$.
If $k\%2=0$, then $(k-1)\%2=1$, $m^{(k-1)}_{u}=0$ for any ${u\in\mathcal N_{G}(v)\cap (V_{G}/V_{G_{S}})}$ (the same as $K$=1), thus $\mathbf h^{(k-1)}_{u}=\mathbf 0$ and $\sum_{u\in\mathcal{N}_{G}(v)\cap (V_{G}/V_{G_{S}})}m^{(k)}_{u}\mathbf W^{(k)}_{2}\\\mathbf h^{(k-1)}_{u}=\mathbf0$. Therefore, we have
\begin{equation}
\begin{aligned}
\nonumber
\mathbf h^{(k)}_{v}&=\textrm{ReLU}(\mathbf W^{(k)}_{1}\mathbf h^{(k-1)}_{v}+\sum_{u\in\mathcal{N}_{G}(v)\cap V_{G_{S}}}m^{(k)}_{u}\mathbf W^{(k)}_{2}\mathbf h^{(k-1)}_{u})\\&=\textrm{ReLU}(\mathbf W^{(k)}_{1}\mathbf t^{(k-1)}_{v}+\sum_{u\in V_{G_{S}}}m^{(k)}_{u}\mathbf W^{(k)}_{2}\mathbf t^{(k-1)}_{u})\\&=\mathbf t^{(k)}_{v}.
\end{aligned}
\end{equation}

Finally, we can conclude that $\forall v\in V_{G_{S}},\mathbf h^{(K)}_{v}=\mathbf t^{(K)}_{v}$.
\end{proof}

\section{Proof of Theorem \ref{theorem1}}
\label{proof-theorem1}

\begin{proof}
From Lemma \ref{lemma1}, the problem is converted to find the assignments of $\mathbf M$ such that $\mathbf h^{(K)}_{v}=\mathbf 0$ for any $v\in V_{G}/V_{G_{S}}$.
Meanwhile, the assignments of $\mathbf M$ should be consistent with that in Lemma \ref{lemma1}. Thus, a simple assignments of $\mathbf M$ is
\begin{equation}
\begin{aligned}
\nonumber
\mathbf M_{k,v}=m^{(k)}_{v}=
\begin{cases}
0&\{v\}\subseteq V_{G}/V_{G_{S}}
\\&\wedge ((\mathcal{N}_{G}(v)\cap V_{G_{S}}\neq\emptyset
\\&\wedge k\%2=1)\vee k=K)\\
1&\{v\}\subseteq V_{G_{S}}\wedge \{k\}\subseteq[K].
\end{cases}
\end{aligned}
\end{equation}
Note that 
\begin{equation}
\begin{aligned}
\nonumber
\mathbf h_{G}&=\textrm{SUM}(\{\{\mathbf h^{(K)}_{v}|v\in V_{G}\}\})
\\&=\textrm{SUM}(\{\{\mathbf h^{(K)}_{v}|v\in V_{G_{S}}\}\}\cup\{\{\mathbf h^{(K)}_{v}|v\in V_{G}/V_{G_{S}}\}\}),
\end{aligned}
\end{equation}
where $\{\{\mathbf h^{(K)}_{v}|v\in V_{G}/V_{G_{S}}\}\}=\{\{\mathbf{0}\}\}$.
According to Lemma 1, we have
\begin{equation}
\begin{aligned}
\nonumber
\mathbf h_{G}&=\textrm{SUM}(\{\{\mathbf h^{(K)}_{v}|v\in V_{G_{S}}\}\})
\\&=\textrm{SUM}(\{\{\mathbf t^{(K)}_{v}|v\in V_{G_{S}}\}\})\\&=\mathbf t_{G_{S}}.
\end{aligned}
\end{equation}
\end{proof}

\begin{figure*}[bp]
\centering
\includegraphics[width=0.55\textwidth]{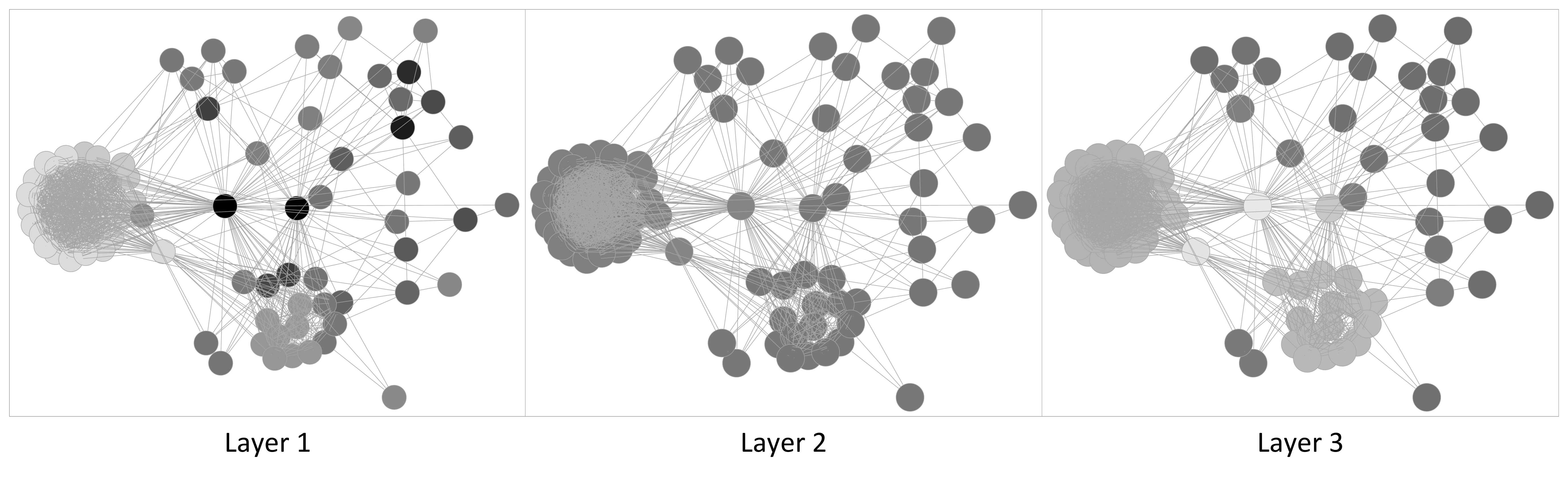}
\includegraphics[width=0.55\textwidth]{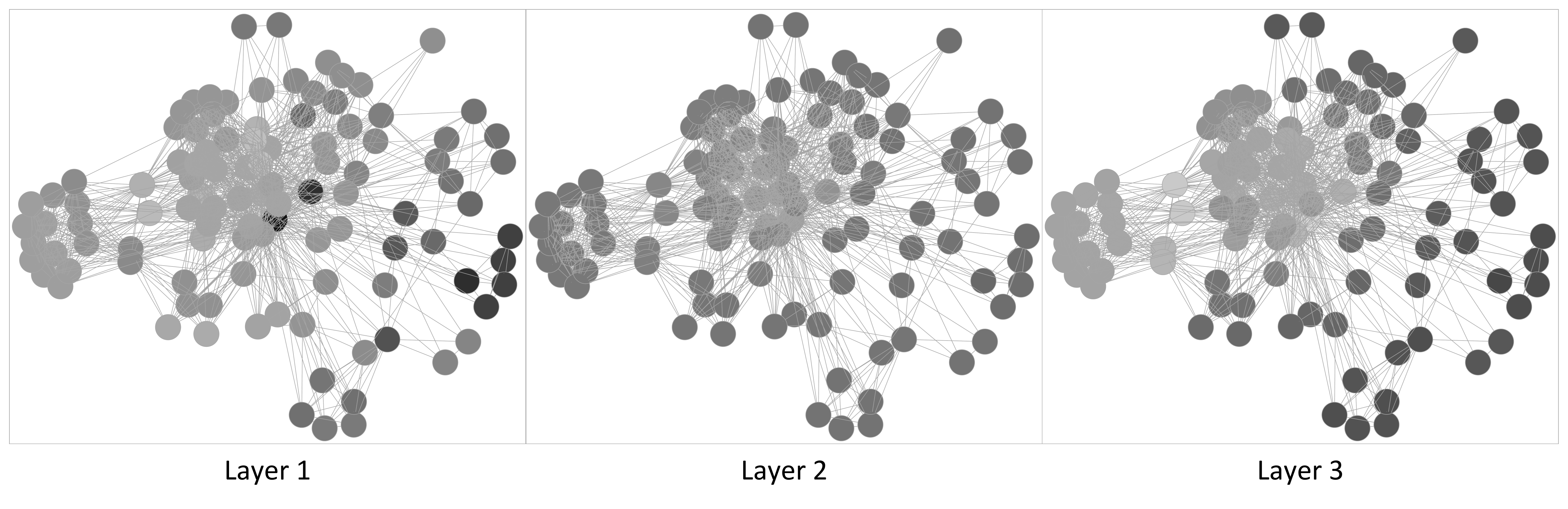}
\caption{Weight distributions of 2 graphs sampled from COLLAB.}
\label{visual-COLLAB}
\end{figure*}
\begin{figure*}[bp]
\centering
\includegraphics[width=0.75\textwidth]{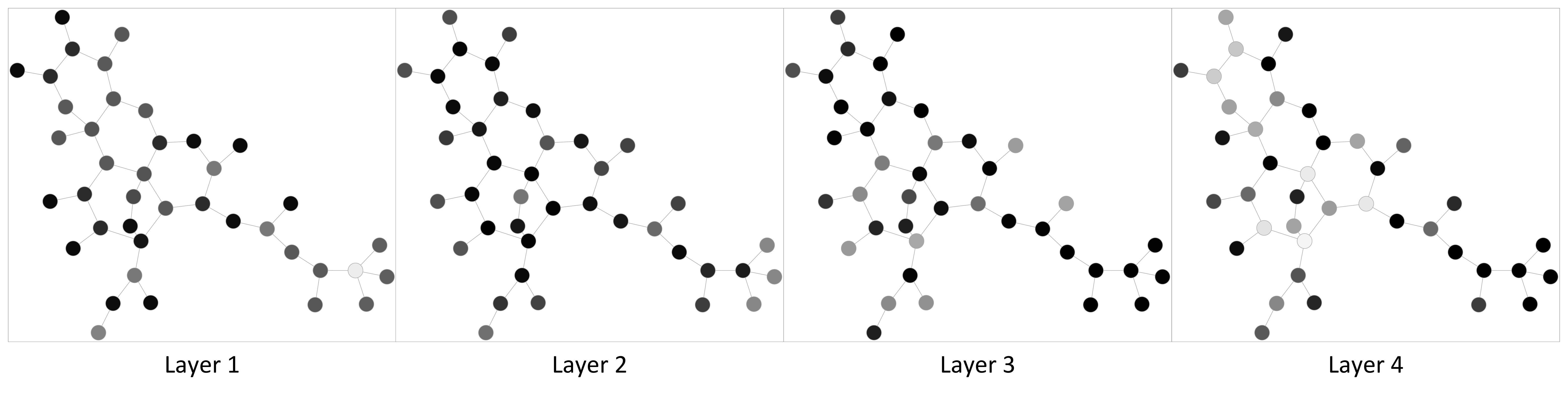}
\includegraphics[width=0.75\textwidth]{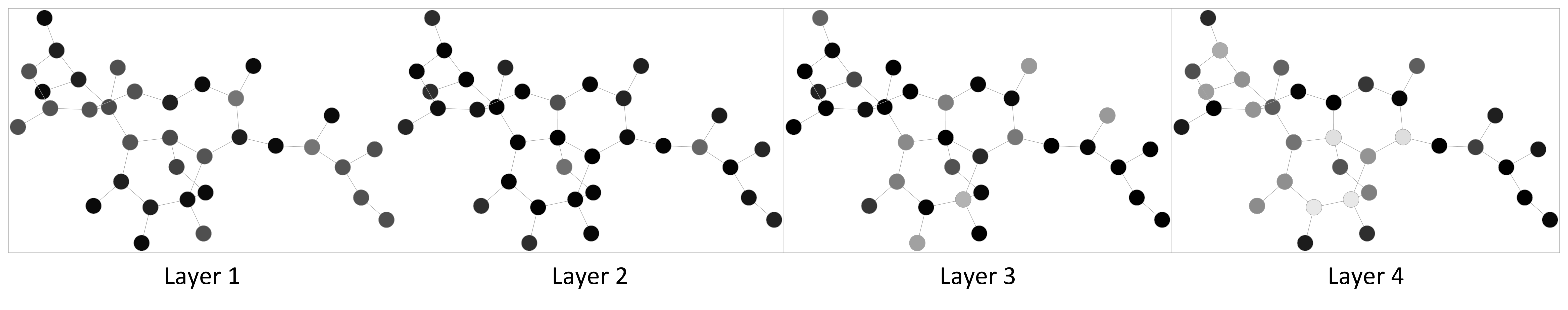}
\caption{Weight distributions of 2 graphs sampled from NCI1.}
\label{visual-NCI1}
\end{figure*}
\begin{figure*}[bp]
\centering
\includegraphics[width=0.9\textwidth]{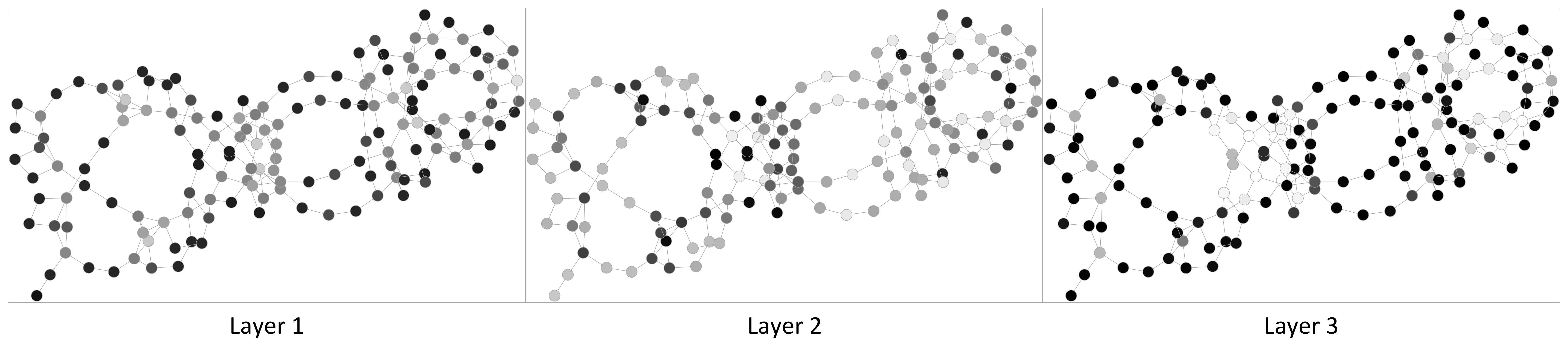}
\includegraphics[width=0.7\textwidth]{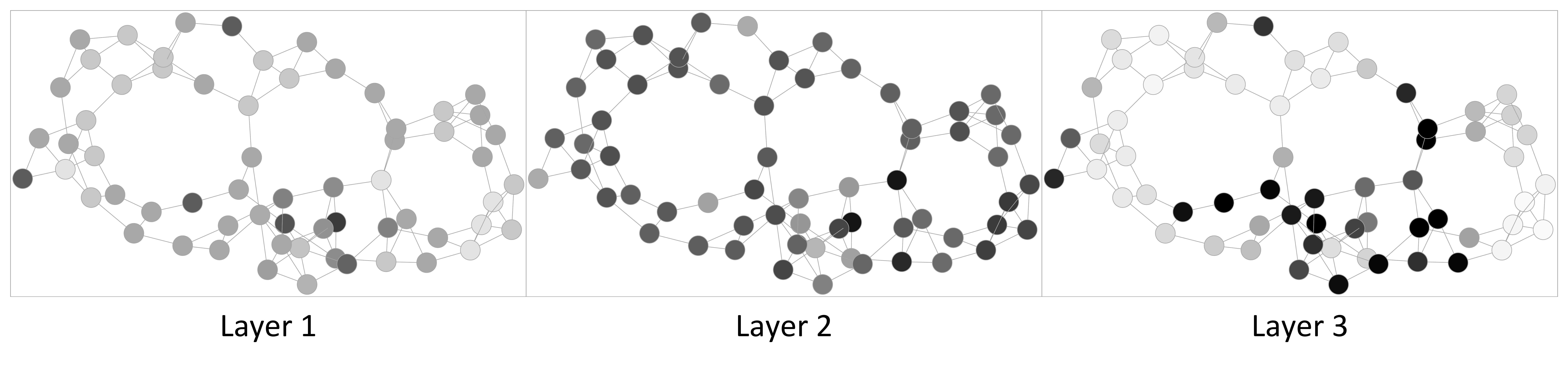}
\caption{Weight distributions of 2 graphs sampled from PROTEINS.}
\label{visual-PROTEINSl}
\end{figure*}

\section{More Visualization Results}
\label{visualizing}

Graphs in COLLAB have dense connections.
For nodes with a large number of neighbors, their values of representation vectors may increase rapidly.
Therefore, in a general GNN, these nodes will dominate the entire graph representation.
Figure \ref{visual-COLLAB} shows that in a soft-mask GNN, nodes with many neighbors learn relative low weights to restrict the fast growth of values of representation vectors.

For small graphs in NCI1 as given in Figure \ref{visual-NCI1}, task-relevant structures should be considered in the entire graph.
The weight distributions in different layers show that the aggregation operation is conducted on most of the nodes, but with different weights.
Graphs in PROTEINS as given in Figure \ref{visual-PROTEINSl} have node attributes, thus the weight distributions are also related to node attributes.
Nodes with 0 weights in the first layer mean that node attributes are not included in the final graph representation, which provides a way to find out task-relevant node attributes.

\end{document}